%% file: main.tex
\documentclass{article}

\usepackage[preprint]{neurips_2025}

\usepackage{amsmath}
\usepackage[table,xcdraw]{xcolor}
\definecolor{highClean}{HTML}{F9B063}
\definecolor{highPoison}{HTML}{BA7FB5}
\definecolor{lowClean}{HTML}{00649D}
\definecolor{lowPoison}{HTML}{B62230}

\definecolor{lightgreen}{HTML}{DCEDC8}
\definecolor{lightred}{RGB}{249,202,202}
\definecolor{lightgray}{RGB}{230,230,230}
\definecolor{citecolor}{HTML}{4D98C9}
\definecolor{linkcolor}{HTML}{c0392b}
\definecolor{boxcolor}{RGB}{194, 213, 247}
\definecolor{lightroyalblue}{HTML}{F6F8FD} 
\definecolor{boxcontentgray}{HTML}{F7F7F7}
\definecolor{boxtitlegray}{HTML}{CCCCCC}
\definecolor{boxbrown}{HTML}{D7CCC8}
\usepackage[hidelinks,breaklinks=true,colorlinks,bookmarks=false,citecolor=citecolor,linkcolor=linkcolor,urlcolor=linkcolor]{hyperref}

\usepackage{graphicx} 
\usepackage[utf8]{inputenc} 
\usepackage[T1]{fontenc}    
\usepackage{url}            
\usepackage{booktabs}       
\usepackage{amsfonts}       
\usepackage{nicefrac}       
\usepackage{microtype}      

\usepackage[capitalize,noabbrev]{cleveref}
\usepackage{enumitem}
\setlist[itemize]{leftmargin=4mm, itemsep=0mm,topsep=0mm, partopsep=0mm}
\usepackage{tabularx}
\usepackage{booktabs} 
\usepackage{multirow}
\usepackage{pifont}
\usepackage{tcolorbox}
\usepackage{graphicx} 
\usepackage{subcaption} 
\long\def\comment#1{}

\def\eg{$e.g.$}

\usepackage{algorithm}
\usepackage{algpseudocode}

\usepackage{amsthm}

\newtheoremstyle{boldhead}
  {3pt}{3pt}
  {\itshape}
  {}
  {\bfseries}
  {.}
  {0.5em}
  {}

\theoremstyle{boldhead}

\newtcbox{\hlgreentab}{on line, rounded corners, box align=base, colback=lightgreen,colframe=white,size=fbox,arc=3pt, before upper=\strut, top=-2pt, bottom=-4pt, left=-2pt, right=-2pt, boxrule=0pt}
\newtcbox{\hlredtab}{on line, box align=base, colback=lightred,colframe=white,size=fbox,arc=3pt, before upper=\strut, top=-2pt, bottom=-4pt, left=-2pt, right=-2pt, boxrule=0pt}
\newtcbox{\hlgraytab}{on line, box align=base, colback=lightgray,colframe=white,size=fbox,arc=3pt, before upper=\strut, top=-2pt, bottom=-4pt, left=-2pt, right=-2pt, boxrule=0pt}

\definecolor{green}{RGB}{0,180,0}
\definecolor{red}{RGB}{180,0,0}

\usepackage{xspace} 

\usepackage{booktabs}
\usepackage{multirow}
\usepackage[normalem]{ulem}
\useunder{\uline}{\ul}{}

\usepackage[export]{adjustbox}
\usepackage{array}
\usepackage{tabularx}
\usepackage{longtable}
\usepackage{amssymb}
\usepackage{tcolorbox}
\usepackage{multirow}

\usepackage{pifont}

\usepackage{caption}
\usepackage{booktabs}
\usepackage{multirow}
\usepackage[table,xcdraw]{xcolor}
\usepackage[normalem]{ulem}
\useunder{\uline}{\ul}{}
\usepackage{wrapfig}
\title{$\mathcal{X}^2$-DFD: A framework for e$\mathcal{X}$plainable and e$\mathcal{X}$tendable Deepfake Detection}

\author{
Yize Chen$^{1*}$, Zhiyuan Yan$^{2*}$,Guangliang Cheng$^{4}$, Kangran Zhao$^{1}$, Siwei Lyu$^{4}$,  Baoyuan Wu$^{1^\dagger}$  \\
  $^1$School of Data Science, \\The Chinese University of Hong Kong, Shenzhen, Guangdong, 518172, P.R. China \\
  $^2$School of Electronic and Computer Engineering, Peking University, P.R. China \\
  $^3$ Department of Computer Science, University of Liverpool, Liverpool, L69 7ZX, UK\\
  $^4$Department of Computer Science and Engineering, University at Buffalo, \\State University of New York, Buffalo, NY, USA
}

\begin{document}

\maketitle

\input{sec/0_abstract}    
\input{sec/1_intro}
\input{sec/2_related}

\input{sec/3_method}
\input{sec/4_experiment}

\input{sec/5_conclusion}

\newpage
{
    \small
    \bibliography{main}
}

\newpage

\end{document}

%% file: sec/0_abstract.tex
\begin{abstract}
This paper proposes \textbf{$\mathcal{X}^2$-DFD}, an \textbf{e$\mathcal{X}$plainable} and \textbf{e$\mathcal{X}$tendable} framework based on multimodal large-language models (MLLMs) for deepfake detection, consisting of three key stages (see \Cref{fig:first}). 
The first stage, {\em Model Feature Assessment}, systematically evaluates the detectability of forgery-related features for the MLLM, generating a prioritized ranking of features based on their intrinsic importance to the model.
The second stage, {\em Explainable Dataset Construction}, consists of two key modules: {\em Strong Feature Strengthening}, which is designed to enhance the model’s existing detection and explanation capabilities by reinforcing its well-learned features, and {\em Weak Feature Supplementing}, which addresses gaps by integrating specific feature detectors (\eg, low-level artifact analyzers) to compensate for the MLLM’s limitations.
The third stage, Fine-tuning and Inference, involves fine-tuning the MLLM on the constructed dataset and deploying it for final detection and explanation.
By integrating these three stages, our approach enhances the MLLM's strengths while supplementing its weaknesses, ultimately improving both the detectability and explainability.
Extensive experiments and ablations, followed by a comprehensive human study, validate the improved performance of our approach compared to the original MLLMs.
More encouragingly, our framework is designed to be plug-and-play, allowing it to seamlessly integrate with future more advanced MLLMs and specific feature detectors, leading to continual improvement and extension to face the challenges of rapidly evolving deepfakes.

\end{abstract}

%% file: sec/1_intro.tex
\section{Introduction}
\label{sec:intro}

Current generative AI technologies have enabled easy manipulation of facial identities, with many applications such as filmmaking and entertainment \citep{deepfake-survey-24}.
However, these technologies can also be misused to create \textit{deepfakes}\footnote{The term ``deepfake" used here refers explicitly to \textbf{face} forgery images or videos. Full (natural) image synthesis is not strictly within our scope.} for malicious purposes, including violating personal privacy, spreading misinformation, and eroding trust in digital media.
Therefore, there is a pressing need to establish a reliable and robust system for detecting deepfakes. In recent years, numerous deepfake detection methods have been proposed \citep{li2018exposing,frequency-detect,mul-attention-detection,li2020face,chen2022self,shiohara2022detecting,yan2023ucf,yan2024transcending}, with the majority focusing on addressing the generalization issue when the manipulation methods between training and testing vary. 
However, these methods typically only output a probability indicating whether a given input is AI-generated, without providing intuitive and convincing explanations behind the prediction. 

\begin{figure}[t]
    \centering
    \includegraphics[width=\linewidth]{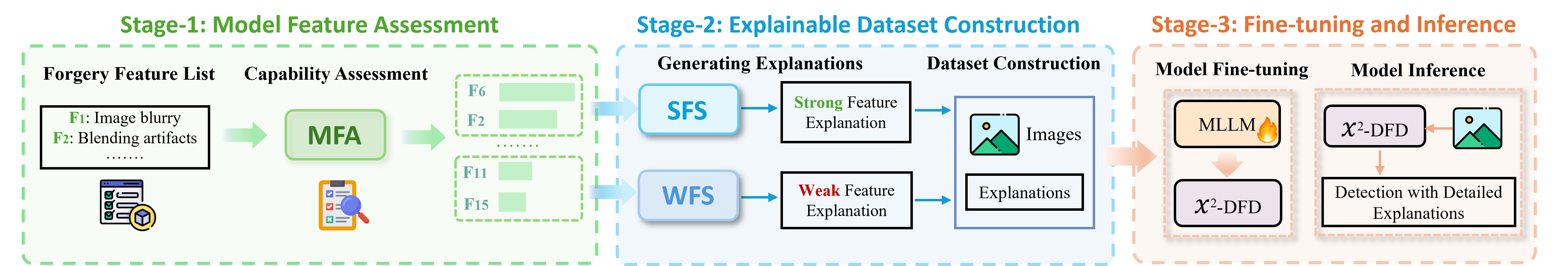}
    \caption{High-level overview of our framework, consisting of three key stages: (1) \textit{Model Feature Assessment (MFA)} evaluates and ranks the forgery-related features (\eg, blending artifacts) to generate a feature set, (2) \textit{Strong Feature Strengthening (SFS)} enhances the model's strong features for improvsed detection and explanation, while \textit{Weak Feature Supplementing (WFS)} leverages \textit{Specific Feature Detector (SFD)} to compensate the model's weak features, and eventually resulting in an explainable dataset, and (3) The MLLM is fine-tuned using the dataset and then used for inference.}
    \vspace{-5pt}
    \label{fig:first}
\end{figure}

Multimodal Large Language Models (MLLMs) have shown remarkable potential in many vision tasks \citep{wu2023multimodal,yan2025gpt,deepfake-survey-24}. 
Given their strong vision-language reasoning capabilities, MLLMs offer a promising avenue for addressing the explainability gap in visual forgery detection. 
Recent studies \citep{jia2024can,shi2024shield,li2024forgerygpt,ye2024loki} have explored this direction by prompting human annotators or LLMs to describe forgery cues from multiple dimensions, which the MLLMs are then trained to detect.
However, these approaches often overlook a key challenge: the \textbf{reliability} of the generated explanations. Due to MLLMs' well-documented tendency to hallucinate, especially under uncertain conditions \citep{bai2024hallucination}, it is \textbf{crucial to ensure that the models relies on their ``familiar'' forgery cues with strong discrimination for detection}. 
Intuitively, not all forgery features are equally useful—some can be effectively leveraged for detection, while others are weakly utilized or ignored altogether.

\begin{wrapfigure}{ht!}{0.5\textwidth}
  \vspace{-0.10in}
  \begin{center}
    \includegraphics[width=1.0\linewidth]{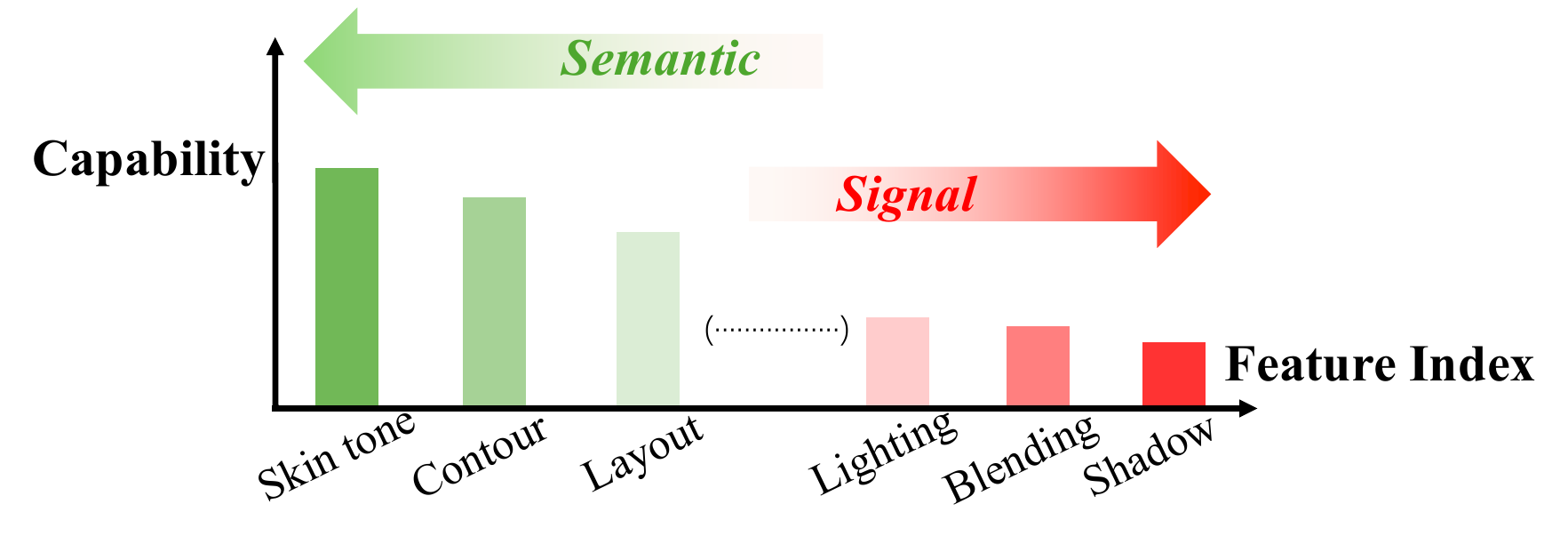}
  \end{center}
  \vspace{-0.15in}
  \caption{The diagram shows that pretrained models (\eg, LLaVa) effectively distinguish real from fake content using \textit{semantic} features (\eg, Skin tone, Contour), but perform poorly with \textit{signal} features (\eg, Blending, Lighting).}
    \label{fig: semantic signal}
  \vspace{-0.15in}
\end{wrapfigure}

To investigate this, we conduct a \textbf{comprehensive analysis of how well pre-trained MLLMs can utilize various forgery-related cues.} As shown in \Cref{fig: semantic signal}, certain cues exhibit strong detection performance (\eg, facial structures and skin tone), whereas others offer limited discriminative value (\eg, blending artifacts and lighting inconsistencies). When a cue is unfamiliar or ineffective for the model, explanations based on it become unreliable. In contrast, cues that align well with the model’s capabilities produce more robust and trustworthy explanations. Therefore, to ensure reliable explanations, it is essential to \textit{explicitly} identify and promote cues that the MLLM can reliably understand and leverage.



Inspired by the above investigations, we propose $\boldsymbol{\mathcal{X}}^2$\textbf{-DFD}, a novel framework that utilizes MLLMs for deepfake detection.
The key idea of our approach is to \textbf{enhance the strengths and supplement the weaknesses} of the original MLLMs. 
Our framework operates through three core stages.
\textbf{First}, the \textit{Model Feature Assessment (MFA)} assesses the intrinsic capability of the original MLLMs in deepfake detection. 
This stage quantifies the discriminative capability of each forgery-related feature, producing a prioritized ranking based on its importance to the model.
\textbf{Second}, the \textit{Strong Feature Strengthening (SFS)} and \textit{Weak Feature Supplementing (WFS)} reinforce strong features and compensate for weak ones, resulting in a more explainable dataset.
\textbf{Third}, we use the created dataset from the second stage to fine-tune the MLLM and then use it for improved detection and explanation.
This integration enables us to leverage the strengths of both MLLMs and Specific Feature Detectors (SFDs) effectively and \textbf{fuse the large and small models adaptively.} 
Encouragingly, the modular-based design of the proposed $\boldsymbol{\mathcal{X}}^2$\textbf{-DFD} framework enables seamless integration with future MLLMs and SFDs as their capabilities evolve.

Our main contributions are threefold. 
\begin{itemize}
    \item \textbf{We systematically assess the intrinsic capabilities of MLLMs for deepfake detection}: To our knowledge, we are the first work to provide an in-depth analysis of MLLMs’ inherent ability in deepfake detection. Our findings reveal that MLLMs exhibit varying discrimination capabilities across different forgery features.
    
    \item \textbf{We enhance MLLMs' explainability by reinforcing their strong features}: Building on their strengths, we fine-tune MLLMs to generate explanations based on their most ``familiar" forgery features, improving both detection accuracy and explainability.

    \item \textbf{We further integrate Specific Feature Detectors to supplement the model's weakness}, For forgery features where MLLMs struggle, we incorporate SFDs to complement their limitations, creating a more robust detection system.
\end{itemize}



%% file: sec/2_related.tex
\vspace{-2pt}
\section{Related Work}

\label{sec:related_work}
\vspace{-2pt}

\paragraph{Conventional Deepfake Detection}
Early detection methods typically focus on performing feature engineering to mine a manual feature such as eye blinking frequency \citep{eye-blink}, warping artifacts \citep{li2018exposing}, headpose \citep{yang2019exposing}, and \textit{etc}.
Recent conventional deepfake detectors mainly focus on dealing with the issue of generalization \citep{yan2023deepfakebench}, where the distribution of training and testing data varies.
Until now, there have developed novel solutions from different directions: constructing pseudo-fake samples to capture the blending clues \citep{li2018exposing,li2020face,shiohara2023blendface,zhao2021learning}, learning spatial-frequency anomalies \citep{gu2022exploiting,frequency-detect,luo2021generalizing,qian2020thinking}, focusing on the ID inconsistency clues between fake and corresponding real \citep{dong2023implicit}, performing disentanglement learning to learn the forgery-related features \citep{yan2023ucf,yang2021learning,fu2025exploring}, performing reconstruction learning to learn the general forgery clues \citep{cao2022end,wang2021representative}, locating the spatial-temporal inconsistency \citep{haliassos2021lips,wang2023altfreezing,zheng2021exploring,yan2024generalizing,zhang2024learning}, and \textit{etc}.
However, these methods can only provide real or fake predictions without providing detailed explanations. The lack of convincing and human-comprehensible explanations might confuse users about why the predictions are deemed fake.

\vspace{-5pt}

\paragraph{Deepfake Detection via Multimodal Large Language Model}
Vision and language are the two important signals for human perception, and visual-language multimodal learning has thus drawn a lot of attention in the AI community. 
Recently, the LLaVA series \citep{liu2024visual,liu2024llava,liu2024improved} have explored a simple and effective approach for visual-language multimodal modeling. 
In the field of deepfake detection, \citep{jia2024can,shi2024shield} have investigated the potential of prompt engineering in face forgery analysis and proposed that existing MLLMs show better explainability than previous conventional deepfake detectors. In addition, \cite{li2024fakebench,foteinopoulou2025hitchhiker,li2024forgerygpt} probed different MLLMs for explainable fake image detection and \cite{li2024fakebench} by presenting a labeled multimodal database for fine-tuning.
More recently, \citep{zhang2024common} proposed using pairs of human-generated visual questions answering (VQA) to construct the fine-tuning dataset, but manually creating detailed annotations can be very costly.
Addressing this limitation, \citep{huang2024ffaa} recently introduced an automated pipeline using GPT-4o \citep{achiam2023gpt} to generate VQA pairs for dataset construction and MLLM training.
However, a new critical question was then raised: \textit{Can MLLMs (\textit{\eg,} LLaVa) fully comprehend the fake clues identified by GPT-4o?} 
We argue that there could remain a ``capability gap" between different MLLMs, particularly between ``annotation generators" (GPT-4o) and ``consumer models" (LLaVA).
This gap exposes two unresolved challenges: (1) systematically analyzing the limitations of MLLM-based detectors in understanding all synthetic forgery clues (\eg, identifying specific detection capabilities they lack) and (2) developing methods to enhance their existing strengths (\eg, semantic consistency analysis) while compensating for weaknesses (\eg, fine-grained artifact recognition).
To our knowledge, most existing works fail to adequately address the two key challenges, leaving a critical void in building more robust and explainable deepfake detection systems.


%% file: sec/3_method.tex
\section{Method}
\label{ref:method}

\begin{figure*}[ht!]
    \centering
    \includegraphics[width=0.96\textwidth]{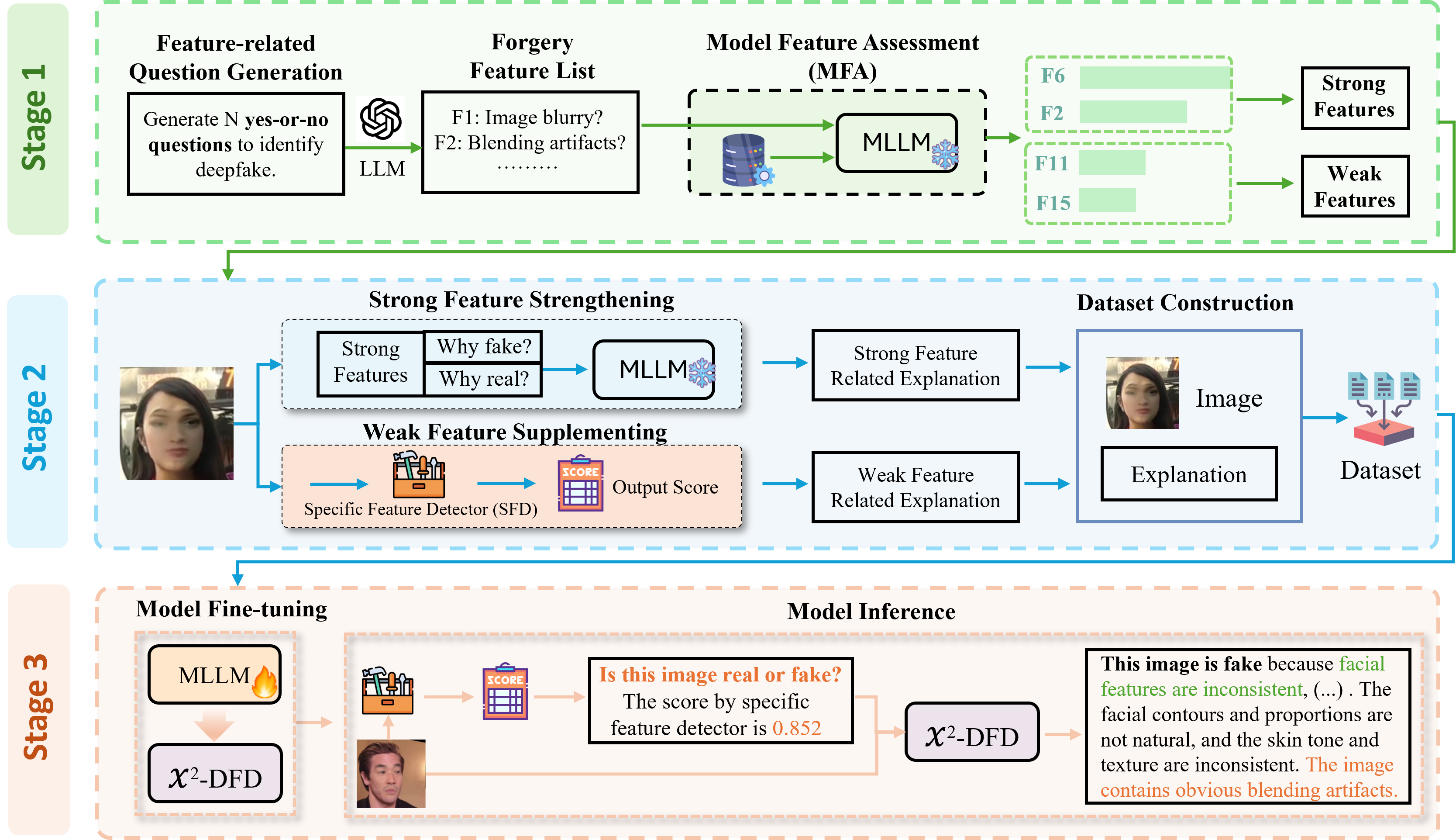}
    \caption{A comprehensive breakdown of the three-stage methodology for \textbf{$\mathcal{X}^2$-DFD}. In \textbf{Stage 1}, an automated procedure for forger-related feature generation, evaluation, and ranking is implemented within the \emph{MFA} (Model Feature Assessment) module. \textbf{Stage 2} incorporates the \emph{SFS} (Strong Feature Strengthening) module, which automates the generation of explanatory annotations for a fine-tuning dataset consisting of real and fake images, leveraging strong features, alongside the \emph{WFS} (Weak Feature Supplementing) module, which employs a specific feature detector to produce explanations for weak features. \textbf{Stage 3} entails model fine-tuning and inference, empowering the model to excel in detection performance and provide precise explanations, utilizing both its proficient strong features and less proficient weak features for improved detection and explanation.}
    \label{fig:MLLM_pipeline}
    \vspace{-10pt}
\end{figure*}

In this work, we propose a general explainable and extendable multimodal framework for deepfake detection, which consists of three key stages: (1) \textit{Model Feature Assessment (MFA)} evaluates and ranks the forgery-related features, (2) \textit{Strong Feature Strengthening (SFS)} enhances the model's strong features and \textit{Weak Feature Supplementing (WFS)} leverages \textit{Specific Feature Detector (SFD)} to compensate the model's weak features, and 
eventually resulting in an explainable dataset, and (3) The MLLM is fine-tuned on this dataset and then used for inference for enhanced deepfake detection and explanations.
In the following content, we will introduce the technical details of these stages.

\subsection{Model Feature Assessment (MFA)}

As depicted in the top row of \Cref{fig:MLLM_pipeline}, the MFA module consists of three sequential stages: \textbf{feature-related question generation, assessment, and ranking}. Each stage plays a crucial role in identifying and prioritizing forgery-related features.

\textit{Step 1: Feature-related Question Generation.}  
For each candidate forgery-related feature, a corresponding question is formulated to assess its presence in an image. Given that these features are not predefined, a Large Language Model (LLM), such as GPT-4o, is leveraged to generate a diverse set of $N_f$ questions, denoted as $ F_i $. These questions are designed to probe key forgery indicators, including facial inconsistencies, unnatural color, and texture mismatches elements critical for deepfake detection.

\textit{Step 2: Model Feature Assessment.}  
Each generated question is paired with an image from the assessment dataset, forming structured prompts for model inference. The Multi-Modal Large Language Model (MLLM) then responds with binary outputs (\texttt{yes} or \texttt{no}), which are aggregated into a confusion matrix to quantify the reliability of each feature. Specifically, for an image $ x_j $ and question $ F_i $, where $i$ represents the index of forgery feature-related question, and $j$ represents index of an image, the MLLM produces:

\begin{equation}
    R_{i,j} = \mathcal{M}_{\text{base}}(F_i, x_j),
\end{equation}
where $ R_{i,j} \in \{\texttt{yes}, \texttt{no}\} $, representing the model's response. This step ensures that the generated questions effectively capture forgery-related discrepancies.

\textit{Step 3: Feature Ranking.}  
To prioritize the most discriminative features, questions are ranked based on their \textit{Balanced Accuracy (BA)}:

\begin{equation}
    \text{BA}_{i} = \frac{1}{2} \left( \frac{\text{TP}_{i}}{\text{TP}_{i} + \text{FN}_{i}} + \frac{\text{TN}_{i}}{\text{TN}_{i} + \text{FP}_{i}} \right),
\end{equation}
where $ \text{TP}_{i}, \text{TN}_{i}, \text{FP}_{i}, \text{FN}_{i} $ denote True Positives, True Negatives, False Positives, and False Negatives, respectively. Due to potential class imbalance between real and fake samples in the dataset, we use the simple and widely adopted \textit{Balanced Accuracy (BA)} metric. This fairly evaluates both classes, aiding effective feature ranking. The ranking identifies the most reliable forgery-related features for discrimination.

Following the automated ranking, \textbf{human verification} is further conducted to ensure the reliability of the identified fake features. This step mitigates potential biases or misinterpretations by the LLM, refining the final selection of discriminative features. Additionally, irrelevant or non-discriminative features are filtered out, with minimal instances of erroneous or unrelated outputs. 


\subsection{Strong Feature Strengthening (SFS)} 
The SFS module constructs datasets by leveraging the strong feature capabilities identified as high-performing by the MFA module. This process comprises two key steps.

\textit{Step 1: Real/Fake Prompts Generation.} 
Leveraging the strong features from the MFA module, we generate specialized prompts to guide the MLLM’s focus during the fine-tuning phase. 
Specifically, we first utilize GPT-4o to summarize these strong features and construct two distinct prompts: one tailored for real images (\( \mathbf{P}_{\text{real}} \)) and another for fake images (\( \mathbf{P}_{\text{fake}} \)). These prompts are formulated as: $\mathbf{P}_{\text{real}} = f(\mathbf{F}_{\text{real}}), \quad \mathbf{P}_{\text{fake}} = f(\mathbf{F}_{\text{fake}}),$
where \( \mathbf{F}_{\text{real}} \) and \( \mathbf{F}_{\text{fake}} \) denote the sets of strong features relevant to real and fake images, respectively. Also, \( f \) represents any LLMs. Here, we employ GPT-4o for implementation.

\textit{Step 2: Fine-tuning Dataset Construction.} 
A fine-tuning dataset $D_{ft}$ comprising VQA-style (visual question answering) pairs, which is constructed by pairing each image with the corresponding (real or fake) prompt. 
Each image is annotated with the specific features it exhibits, and the standardized prompt \( \mathbf{P}_{\texttt{fixed}} \) is defined as: $\mathbf{P}_{\texttt{fixed}} = \text{``Is this image real or fake?"}$
The model's response is structured to begin with a definitive statement---``\texttt{This image is real/fake}"---followed by an explanation based on the identified features. 
Formally, the final answer is represented as: $\mathbf{A}_{\texttt{final}} = ``\texttt{This image is real/fake}" + \mathbf{A}_{\texttt{real/fake}}$.
Consequently, each VQA-style pair of the fine-tuning dataset $D_{ft}$ is formalized as: $\mathbf{VQA} = (\texttt{Image}, \mathbf{P}_{\texttt{fixed}}, \mathbf{A}_{\texttt{final}}).$


\subsection{Weak Feature Supplementing (WFS)} 
The WFS module construct datasets by integrating specific feature detectors, which are specialized in detecting features where the MLLM shows weakness.
This module follows two steps:

\textit{Step 1: Specific Feature Detector Invocation.}
For features that the MLLM identifies as weak, we deploy an specialized deepfake detector (\eg, a blending-based detector \citep{lin2024CDFA}). This specific feature detector processes the input image and generates a prediction.
Note that we also employ other SFDs for implementation, and we provide an in-depth analysis for this in the \Cref{section: WFS analysis}.
Specifically, when utilizing a blending detector as an instance of SFD, a blending score \( s \) is produced: $s = \sigma (\text{BlendDetector}(x))$,
where \( x \) denotes the input image, and \(\sigma \) denotes the sigmoid function that transforms the logits output of the $\text{BlendDetector}$ into the 0-1 range.

\textit{Step 2: Integration of Specific Feature Detection Results into the Fine-tuning Dataset.} 

The blending score \( s \) obtained from the specific detector is incorporated into the fine-tuning dataset by appending it to the existing prompts. This is done by adding a statement such as: ``\texttt{And the blending feature score of content is: \underline{s}}" Additionally, based on the score, a corresponding response aligned with the probability is included, specifically in the Fine-tuning Dataset Construction section of the SFS. This integration ensures that the MLLM benefits from both its intrinsic detection capabilities and the specialized insights provided by the SFD.

\subsection{Model Finetune and Inference} 
\label{finetune and infer}

After obtaining the constructed dataset, the following steps involve fine-tuning and inference.
The stage following two steps:

\textit{Step 1: MLLM Fine-tuning.} 
The initial MLLM is fine-tuned using the dataset \( D_{ft} \). This process adjusts the \textit{projector} to accurately link image artifacts with corresponding fake labels. Additionally, Low-Rank Adaptation (LoRA) \citep{hu2021lora} is applied to selectively update a subset of the model’s parameters, enhancing its focus on deepfake-specific features while preserving overall model integrity. This fine-tuning can be expressed as: 
\[ \mathcal{M}_{\text{base}} \xrightarrow[]{D_{ft}} \mathcal{M}_{\text{fine-tuned}}, \]
where \( \mathcal{M}_{\text{fine-tuned}} \) represents the enhanced MLLM with superior deepfake detection capabilities.

\textit{Step 2: Integration of Specific Feature Detection into Inference Prompts.} 
Generally, during the inference, the SFD detector’s blending score \( s \) is incorporated into the MLLM’s prompt-based reasoning process. The final output of the model is structured, to begin with a definitive statement: \texttt{"This image is real/fake"}, followed by reasoning based on identified visual features. Based on the blending score \( s \), the model appends a descriptive statement: $\mathbf{A}_{\texttt{final}}$ = \texttt{"This image is real/fake"} (binary results) + $\mathbf{A}_{\texttt{real/fake}}$ (explanations) + \texttt{"And this image contains obvious/minimal blending artifacts"} (clues from SFD). The model acquires this response pattern through training. This approach ensures that the MLLM effectively leverages SFDs to enhance its detection performance, particularly for features where it initially demonstrated weakness.


%% file: sec/4_experiment.tex
\section{Experiment}
\label{ref:experiment}

\subsection{Experimental Setup}

\paragraph{Datasets.}

We evaluate our proposed method on a diverse set of widely-used deepfake detection datasets, including the Deepfake Detection Challenge (DFDC)~\citep{dfdc}, its preview version (DFDCP)~\citep{dfdcp}, DeepfakeDetection (DFD)~\citep{dfd}, Celeb-DF-v2 (CDF-v2)~\citep{li2019celeb}, FaceForensics++ (FF++)~\citep{rossler2019faceforensics++} (c23 version for training), DFo~\citep{jiang2020deeperforensics}, WildDeepfake (WDF)~\citep{zi2020wilddeepfake}, FFIW~\citep{zhou2021face}, and the newly released DF40 dataset~\citep{yan2024df40}, which incorporates state-of-the-art forgery techniques such as Facedancer~\citep{rosberg2023facedancer}, FSGAN~\citep{nirkin2019fsgan}, inSwap~\citep{smith2020roop}, e4s~\citep{li2023e4s}, Simswap~\citep{chen2020simswap}, and Uniface~\citep{zhou2023uniface}. In line with the standard deepfake benchmark \citep{yan2023deepfakebench}, we use the c23 version of FF++ for training and other datasets for testing.


\paragraph{Evaluation Metrics.}
We assess the performance of our model in terms of both detection performance and explanation quality. \textbf{For detection}, we adopt the Area Under the Curve (AUC) as the primary metric to evaluate the model’s ability to distinguish real from fake content across entire datasets, reporting both frame-level and video-level AUC scores. Additional metrics, including Accuracy (Acc.), Equal Error Rate (EER), and Average Precision (AP), are also provided for a comprehensive analysis. 
\textbf{For explanation}, we follow~\cite{zhang2024common} by using human-annotated data to measure text similarity between model-generated explanations and human-labeled ground truth, employing standard metrics such as \textit{BLEU}~\cite{papineni2002bleu}, \textit{CIDEr}~\cite{vedantam2015cider}, \textit{ROUGE-L}~\cite{lin2004rouge}, \textit{METEOR}~\cite{denkowski2014meteor}, and \textit{SPICE}~\cite{anderson2016spice}. Beyond text similarity, we engage human evaluators and GPT-4o to assess the quality of explanations regarding forgery content, following prior studies~\cite{xu2024fakeshield,foteinopoulou2025hitchhiker}. Evaluators rate the explanations on a scale from 0 (very poor) to 5 (excellent), ensuring a robust qualitative evaluation.

\vspace{-5pt}
\paragraph{Implementation Details.} 
We initialize our model with the LLaVA-base weights and fine-tune the LLaVA model~\citep{liu2024visual} using its official implementation codebase. For the specific feature detectors (SFD), we adopt a blending-based approach as proposed in~\citep{lin2024CDFA}. Training is performed on a single NVIDIA 4090 GPU for 3 epochs, with a learning rate of $2 \times 10^{-5}$ in two layer mlp projector and $2 \times 10^{-4}$ in others, a rank of 16, and an alpha value set conventionally to twice the rank at 32. We use a batch size of 4, a gradient accumulation step of 1, and a warmup ratio of 0.03 to stabilize training. 

\input{tables/corss_datasets}



\subsection{Generalizability Evaluation}

Following the common settings of previous works~\citep{yan2024transcending,cheng2024can}, we first compare our method with 33 SOTA detectors (including both conventional and multimodal-based detectors) via \textbf{cross-dataset evaluations} (see \Cref{tab:Protocol-1})
The results of other compared baselines are mainly cited from their original papers.
Our approach excels across both frame-level and video-level evaluations, maintaining superior results when compared to other methods. The table clearly highlights our method's capability to generalize and consistently achieve higher detection performance at the frame level and video level, respectively. 

\subsection{Explainability Evaluation}
\paragraph{Annotated Explainability Evaluation.}
We assess the performance of our model using the DD-VQA~\cite{zhang2024common} test dataset, which incorporates human-annotated data from FF++~\cite{rossler2019faceforensics++}. The evaluation employs a suite of metrics, including \textit{BLEU}~\cite{papineni2002bleu}, \textit{CIDEr}~\cite{vedantam2015cider}, \textit{ROUGE-L}~\cite{lin2004rouge}, \textit{METEOR}~\cite{denkowski2014meteor}, and \textit{SPICE}~\cite{anderson2016spice}, to quantify the alignment between our model's responses and human-annotated ground truth. The MLLMs assessed for explanation quality include LLaVA~\citep{liu2024llava}, Llama3.2V~\citep{dubey2024llama}, Qwen2VL~\citep{wang2024qwen2} and GPT4o~\cite{achiam2023gpt}. The models use the same prompt to generate explainable outputs. To ensure a fair comparison, particularly given GPT-4o's tendency to refuse responses with the same prompt, we adopt a prompting strategy from~\cite{jia2024can}. This leads GPT-4o to generate shorter responses, resulting in lower scores. The evaluation results are summarized in \Cref{tab:combined_evaluation} \textit{(Annotated Explainability Evaluation)}. Due to the DD-VQA only annotating the artifact in some specific parts (\eg, {\em nose, eyes}), GPT-4o and human experts are required to evaluate both annotated and unannotated scenarios, ensuring a thorough assessment of model explainability across different scenarios.

\input{tables/explain}
\paragraph{Unannotated Explainability Evaluation.}
To evaluate unannotated explainability, we build on insights from prior work~\cite{xu2024fakeshield, foteinopoulou2025hitchhiker} and utilize both human evaluators and GPT-4o to assess model performance across three key dimensions: (1) detection ability, (2) reasonableness of explanations, and (3) level of detail. Each dimension is scored on a scale from 0 to 5. The evaluation results are summarized in \Cref{tab:combined_evaluation} \textit{(Unannotated Explainability Evaluation)}. The results demonstrate that our model achieves strong performance in both human-eval and GPT-eval. Additional details include the experiment of the setting of human study, Graphical User Interface (GUI) of human study, and the evaluation prompt of GPT4o can be found in Appendix.

\vspace{-3pt}
\section{Ablation Study and Analysis}
\vspace{-5mm}
Here, we address several \textbf{key research questions} through ablation studies and in-depth analysis.

\colorbox{gray!20}{\strut \textit{Question 1: Why is fine-tuning MLLMs with their strong features more effective than using all?}}

To enhance the reasoning and detection capabilities of MLLMs, we introduce the Strong Feature Strengthening (SFS) module. In this module, we focus on selectively amplifying the most discriminative forgery-related features—referred to as strong features. These strong features are identified through Multidimensional Feature Attribution (MFA), which ranks features based on their importance scores across different modalities and samples.

\begin{wraptable}{r}{0.65\textwidth}
\vspace{-0.5cm}
\centering
\captionsetup{font=small}
\caption{Results comparing AUC, AP, and EER using all features versus strong features, both enhanced by SFS module.}
\renewcommand\arraystretch{1.1}
\resizebox{0.65\textwidth}{!}{%
\setlength{\tabcolsep}{2pt}
\begin{tabular}{l|ccc|ccc|ccc|ccc}
\toprule
\multirow{2}{*}{Feature Selection} & \multicolumn{3}{c|}{CDF} & \multicolumn{3}{c|}{DFDC} & \multicolumn{3}{c|}{Uniface} & \multicolumn{3}{c}{Average} \\ \cmidrule{2-13} 
 &
  \multicolumn{1}{c|}{AUC} &
  \multicolumn{1}{c|}{AP} &
  EER &
  \multicolumn{1}{c|}{AUC} &
  \multicolumn{1}{c|}{AP} &
  EER &
  \multicolumn{1}{c|}{AUC} &
  \multicolumn{1}{c|}{AP} &
  EER &
  \multicolumn{1}{c|}{AUC} &
  \multicolumn{1}{c|}{AP} &
  EER \\ \midrule
All Features &
  \multicolumn{1}{c|}{79.0} &
  \multicolumn{1}{c|}{88.3} &
  28.9 &
  \multicolumn{1}{c|}{77.8} &
  \multicolumn{1}{c|}{81.9} &
  28.9 &
  \multicolumn{1}{c|}{82.3} &
  \multicolumn{1}{c|}{84.8} &
  25.2 &
  \multicolumn{1}{c|}{79.7} &
  \multicolumn{1}{c|}{85.0} &
  27.7 \\
Strong Features &
  \multicolumn{1}{c|}{83.2} &
  \multicolumn{1}{c|}{90.5} &
  24.6 &
  \multicolumn{1}{c|}{79.2} &
  \multicolumn{1}{c|}{82.1} &
  27.6 &
  \multicolumn{1}{c|}{84.5} &
  \multicolumn{1}{c|}{86.2} &
  22.4 &
  \multicolumn{1}{c|}{82.3} &
  \multicolumn{1}{c|}{86.3} &
  24.9 \\ \midrule
\end{tabular}
}
\vspace{-4mm}
\label{tab:all or select}
\end{wraptable}

This design leads us to a critical question: \textit{Are these strong features truly more effective for improving model performance than using the full set of features?} 
To answer this, we compare two strategies: (1) enhancing all features in the LLM-generated feature list and (2) enhancing only the strong features selected via MFA. As shown in \Cref{tab:all or select}, the latter consistently outperforms the former across all datasets. These results confirm that selectively strengthening the most discriminative features not only improves the model’s performance but also yields more reliable model explanations.

\colorbox{gray!20}{\strut \textit{Question 2: How to ensure SFS module works when it should, and stays silent when it shouldn’t?}}

To further improve model generalization and interpretability, we extend our framework with two new components: the Specific Feature Detection (SFD) module and the Weak Feature Supplementing (WFS) module. While the earlier SFS module focuses on enhancing strong, highly discriminative features, it may overlook subtle patterns critical for certain forgery types. To address this, WFS is designed to teach the model how to leverage weak features provided by SFD—features that are otherwise hard for MLLMs to interpret directly.

\begin{wraptable}{r}{0.49\textwidth}
\vspace{-0.5cm}
\centering
\captionsetup{font=small}
\caption{Effect of excluding supplementary features during training (WFS) and including them at inference (SFD infer) on model performance.}
\renewcommand\arraystretch{1.1}
\resizebox{0.49\textwidth}{!}{%
\setlength{\tabcolsep}{2pt}

\begin{tabular}{l|ccccc}
\toprule
Varient                 & CDF    & DFD    & DFDC   & DFDCP  & Avg.    \\ \midrule
WFS \ding{55}  SFD infer \ding{55}  & 83.2 & 91.4 & 79.2 & 82.0 & 84.0 \\
WFS \ding{55} SFD infer \ding{51} & 81.7 & 90.6 & 79.1 & 81.3 & 83.2 \\
WFS \ding{51} SFD infer \ding{51} & \textbf{90.4} & \textbf{92.3} & \textbf{83.7} & \textbf{87.3} & \textbf{88.4} \\ \bottomrule
\end{tabular}
}
\vspace{-0.2cm}
\label{tab: SFD and WFS}
\end{wraptable}

To investigate whether this combination yields synergistic benefits, we compare three variants: (1) baseline without WFS and SFD at inference, (2) enabling SFD only during inference (without WFS), and (3) enabling both WFS and SFD. As shown in \Cref{tab: SFD and WFS}, the model achieves the best performance when WFS is present, demonstrating that \textbf{SFD’s weak signals become more useful once the model has learned how to utilize them through WFS}. Without WFS, simply adding SFD at inference may not help—and can even lead to degradation—indicating that "1+1" only becomes greater than 2 when weak features are integrated structurally during training.

\begin{wraptable}{r}{0.65\textwidth}
\vspace{-0.3cm}
\centering
\captionsetup{font=small}
\caption{Comparison of AUC performance for models trained on FF++ alone versus FF++ with SRI, evaluated on other datasets.}
\vspace{-4pt}
\renewcommand\arraystretch{1.1}
\resizebox{0.65\textwidth}{!}{%
\setlength{\tabcolsep}{2pt}
\begin{tabular}{l|ccccccccl}
\toprule
Train Data           & CDF           & DFDCP         & DFDC          & DFD           & Uniface       & Fsgan         & Inswap        & Simswap       & Avg. \\ \midrule
FF++ \ding{51} SRI \ding{55}       & 90.4          & 87.3 & 83.7          & 92.3          & 85.5          & \textbf{91.1}          & \textbf{81.2}         & 85.1          & 87.1    \\
FF++ \ding{51} SRI \ding{51} & \textbf{91.5} & \textbf{89.3}          & \textbf{83.9} & \textbf{92.7} & \textbf{87.4} & 89.9 & 81.0 & \textbf{86.1} &  \textbf{87.7}   \\ \bottomrule
\end{tabular}
}
\vspace{-2pt}
\vspace{-0.2cm}
\label{tab: with SRI}
\end{wraptable}
\vspace{-2pt}

Beyond synergy, it is \textbf{crucial to ensure that the introduction of SFD does not interfere in scenarios where its cues are irrelevant}. For example, blending-based detectors may provide limited value on datasets like SRI, which contain no blending traces. We test this by introducing SRI as a training set and comparing performance. As shown in \Cref{tab: with SRI}, the model not only maintains its effectiveness but even improves, suggesting that \textbf{when SFD signals are weak or absent, the model naturally downplays them}. This demonstrates that our design is adaptive—SFD helps when it can, and steps aside when it should.

Overall, our framework achieves both \textbf{synergistic improvement and non-intrusive integration}: \textit{WFS enables the model to benefit from weak features without forcing reliance, and SFD contributes only when its signals are relevant}.

\colorbox{gray!20}{\strut \textit{Question 3: How can we generate the most suitable set of N forgery-related questions in MFA?}}

The Model Feature Assessment (MFA) module evaluates the model’s discriminative ability by asking it to answer a curated set of N forgery-related questions. \textit{A key challenge here is: how to generate the most suitable questions that effectively probe the model’s understanding of diverse forgery cues.}
To explore this, we compare different question-generation strategies:
\textbf{(1)} human-written features based on expert knowledge \citep{zhang2024common},
\textbf{(2)} features automatically generated by large language models (LLMs), including Claude3.5-Sonnet \citep{claude3} and GPT-4o \citep{achiam2023gpt}.

\begin{wraptable}{r}{0.65\textwidth}
\vspace{-0.5cm}
\centering
\captionsetup{font=small}
\caption{Comparison between LLMs and human annotators for generating N forgery-related questions for MFA.}
\renewcommand\arraystretch{1.1}
\resizebox{0.65\textwidth}{!}{%
\setlength{\tabcolsep}{2pt}
\begin{tabular}{l|c|c|c|c|c|c|c|c}
\toprule
Variant                              & CDF  & DFDCP & DFDC & DFD  & Uniface & Fsgan & Simswap & Avg  \\ \midrule
Human Writing \citep{zhang2024common} & 89.1 & 89.7  & 83.6 & 92.5 & 82.3    & 89.1  & 87.0    & 87.6 \\
Claude3.5-Sonnet \cite{claude3}       & 90.1 & 88.5  & 83.5 & 93.0 & 84.9    & 90.0  & 85.6    & 87.9 \\
GPT4o \cite{achiam2023gpt}            & 90.3 & 89.7  & 83.5 & 92.5 & 85.2    & 89.9  & 84.9    & 87.8 \\ \hline
\end{tabular}
}
\label{tab: MFA ablation}
\end{wraptable}

As shown in \Cref{tab: MFA ablation}, questions generated by LLMs slightly outperform those crafted by humans in terms of detection performance across multiple datasets. This suggests that LLMs can capture a broader and potentially more nuanced range of forgery-related features, possibly including cues overlooked by human experts.
However, LLM-generated questions are not always ideal—they may sometimes be generic, redundant, or irrelevant (e.g., mistakenly treating "Photoshop traces" as core deepfake features). On the other hand, although human-designed features may be narrower in scope, they offer higher precision and domain relevance, leading to robust results.
To balance these strengths and weaknesses, we adopt a \textbf{hybrid strategy}:
\textbf{(1)} Use an LLM to generate a diverse pool of forgery-related questions.
\textbf{(2)} Rank them by relevance scores.
\textbf{(3)} Apply human verification to filter out irrelevant or low-quality questions.

\vspace{-5pt}

%% file: tables/corss_datasets.tex
\begin{table*}[ht!]
\centering
\captionsetup{font=small}
\caption{Cross-dataset evaluations with 33 existing detectors. The top two results are highlighted, with the best in \textbf{bold} and the second-best \underline{underlined}. `*' indicates our reproductions based on the pre-trained checkpoints released by the authors, and `$\dagger$' refers to the MLLM-based detectors, which can also output explanations.}
\label{tab:Protocol-1}
\renewcommand\arraystretch{1.1}
\resizebox{\textwidth}{!}{%
\setlength{\tabcolsep}{2pt}
\begin{tabular}{lccccc|lccccc}
\toprule
\multicolumn{6}{c|}{\textbf{Frame-Level AUC}} &
  \multicolumn{6}{c}{\textbf{Video-Level AUC}} \\ \midrule
\multicolumn{1}{c|}{\textbf{Method}} &
  \textbf{CDF} &
  \textbf{DFDCP} &
  \textbf{DFDC} &
  \textbf{DFD} &
  \textbf{Avg.} &
  \multicolumn{1}{c|}{\textbf{Method}} &
  \textbf{CDF} &
  \textbf{DFDCP} &
  \textbf{DFDC} &
  \textbf{DFD} &
  \textbf{Avg.} \\ \midrule
\multicolumn{1}{l|}{Xception \citep{chollet2017xception}} &
  73.7 &
  73.7 &
  70.8 &
  81.6 &
  75.0 &
  \multicolumn{1}{l|}{Xception \citep{chollet2017xception}} &
  81.6 &
  74.2 &
  73.2 &
  89.6 &
  79.7 \\
\multicolumn{1}{l|}{FWA \citep{li2018exposing}} &
  66.8 &
  63.7 &
  61.3 &
  74.0 &
  66.5 &
  \multicolumn{1}{l|}{PCL+I2G \citep{zhao2021learning}} &
  90.0 &
  74.4 &
  67.5 &
  -- &
  -- \\
\multicolumn{1}{l|}{Efficient-b4 \citep{tan2019efficientnet}} &
  74.9 &
  72.8 &
  69.6 &
  81.5 &
  74.7 &
  \multicolumn{1}{l|}{LipForensics \citep{haliassos2021lips}} &
  82.4 &
  -- &
  73.5 &
  -- &
  -- \\
\multicolumn{1}{l|}{Face X-ray \citep{li2020face}} &
  67.9 &
  69.4 &
  63.3 &
  76.7 &
  69.3 &
  \multicolumn{1}{l|}{FTCN \citep{zheng2021exploring}} &
  86.9 &
  74.0 &
  71.0 &
  94.4 &
  81.6 \\
\multicolumn{1}{l|}{F3-Net \citep{qian2020thinking}} &
  77.0 &
  77.2 &
  72.8 &
  82.3 &
  77.3 &
  \multicolumn{1}{l|}{ViT-B (CLIP) \citep{dosovitskiy2020image}} &
  88.4 &
  82.5 &
  76.1 &
  90.0 &
  84.3 \\
\multicolumn{1}{l|}{SPSL \citep{liu2021spatial}} &
  76.5 &
  74.1 &
  70.1 &
  81.2 &
  75.5 &
  \multicolumn{1}{l|}{CORE \citep{ni2022core}} &
  80.9 &
  72.0 &
  72.1 &
  88.2 &
  78.3 \\
\multicolumn{1}{l|}{SRM \citep{luo2021generalizing}} &
  75.5 &
  74.1 &
  70.0 &
  81.2 &
  75.2 &
  \multicolumn{1}{l|}{SBI* \citep{shiohara2022detecting}} &
  90.6 &
  87.7 &
  75.2 &
  88.2 &
  85.4 \\
\multicolumn{1}{l|}{ViT-B (IN21k) \citep{radford2021learning}} &
  75.0 &
  75.6 &
  73.4 &
  86.4 &
  77.6 &
  \multicolumn{1}{l|}{UIA-ViT \citep{zhuang2022uia}} &
  82.4 &
  75.8 &
  -- &
  94.7 &
  -- \\
\multicolumn{1}{l|}{ViT-B (CLIP) \citep{dosovitskiy2020image}} &
  81.7 &
  80.2 &
  73.5 &
  86.6 &
  80.5 &
  \multicolumn{1}{l|}{SLADD* \citep{chen2022self}} &
  79.7 &
  -- &
  77.2 &
  -- &
  -- \\
\multicolumn{1}{l|}{RECCE \citep{cao2022end}} &
  73.2 &
  74.2 &
  71.3 &
  81.8 &
  75.1 &
  \multicolumn{1}{l|}{DCL \citep{sun2022dual}} &
  88.2 &
  76.9 &
  75.0 &
  92.1 &
  83.1 \\
\multicolumn{1}{l|}{IID \citep{huang2023implicit}} &
  83.8 &
  81.2 &
  -- &
  -- &
  -- &
  \multicolumn{1}{l|}{SeeABLE \citep{larue2023seeable}} &
  87.3 &
   86.3 &
  75.9 &
  -- &
  -- \\
\multicolumn{1}{l|}{ICT \citep{dong2022protecting}} &
  85.7 &
  -- &
  -- &
  84.1 &
  -- &
  \multicolumn{1}{l|}{CFM \citep{luo2023beyond}} &
  89.7 &
  80.2 &
  70.6 &
  95.2 &
  83.9 \\
\multicolumn{1}{l|}{LSDA \citep{yan2024transcending}} &
  83.0 &
  81.5 &
  73.6 &
   88.0 &
   81.5 &
  \multicolumn{1}{l|}{UCF \citep{yan2023ucf}} &
  83.7 &
  74.2&
  77.0 &
  86.7 &
  80.4 \\
\multicolumn{1}{l|}{VLFFD$\dagger$ \citep{sun2025towards}} &
  83.2 &
  83.2 &
  -- &
  \textbf{94.8} &
  -- &
  \multicolumn{1}{l|}{LSDA \citep{yan2024transcending}} &
  89.8 &
  81.2 &
  73.5 &
  95.6 &
  85.0 \\ 
  \multicolumn{1}{l|}{FFAA$\dagger$ \citep{huang2024ffaa}} &
 --&
 --&
   74.0&
   92.0&
   -- &
  \multicolumn{1}{l|}{AltFreeing \citep{wang2023altfreezing} } &
  89.5&
  -- &
  -- &
  -- &
   -- \\ 
\multicolumn{1}{l|}{RepDFD$\dagger$ \citep{lin2025standing} } &
  80.0 &
  \textbf{90.6} &
  77.3 &
   -- &
  -- &
  \multicolumn{1}{l|}{TALL-Swin \citep{xu2023tall} } &
   90.8&
     --&
   76.8&
 --&
 --\\
   \multicolumn{1}{l|}{MFCLIP $\dagger$ \citep{lin2025standing} } &
    83.5&
   86.1&
  -- &
 --&
  -- &
  \multicolumn{1}{l|}{StyleDFD \citep{choi2024exploiting} } &
   89.0&
    -- &
   -- &
   {\ul 96.1}&
 --\\
\multicolumn{1}{l|}{KFD-VLM $\dagger$ \citep{yu2025unlocking} } &
    89.9&
  86.7 &
  -- &
   92.3&
  -- &
  \multicolumn{1}{l|}{NACO \citep{zhang2024learning} } &
   89.5&
 --&
   76.7&
 --&
 --\\
   \midrule
\multicolumn{1}{l|}{
\textbf{$\mathcal{X}^2$-DFD (7B)}} &
  { \ul 90.4} &
  87.3 &
  \textbf{83.7} &
  92.3 &
  {\ul 88.4} &
  \multicolumn{1}{l|}{
   \textbf{$\mathcal{X}^2$-DFD (7B)}} &
  {\ul 95.4} &
  {\ul 89.3} &
  \textbf{86.0} &
  95.8 &
  {\ul 91.6} \\
\multicolumn{1}{l|}{\textbf{$\mathcal{X}^2$-DFD (13B)}
} &
  \textbf{91.3} &
  {\ul 90.3} &
  {\ul 83.4} &
  {\ul 92.5} &
  \textbf{89.4} &
  \multicolumn{1}{l|}{\textbf{$\mathcal{X}^2$-DFD (13B)}} &
    \textbf{95.7} &
  \textbf{91.0} &
  {\ul 85.7} &
  \textbf{96.1} &
  \textbf{92.1}  \\
  \bottomrule
\end{tabular}}
\vspace{-5mm}
\end{table*}

%% file: tables/explain.tex
\begin{table*}[h!]
\centering
\captionsetup{font=small}
\caption{Explainability evaluation across annotated and unannotated settings, comparing five  scores for annotated explainability, alongside human and GPT-4o evaluations (scored 0-5) for unannotated explainability. The best result per metric is highlighted in \textbf{bold}.}
\label{tab:combined_evaluation}
\renewcommand\arraystretch{1.1}
\resizebox{\textwidth}{!}{%
\begin{tabular}{l|cccccc|ccc}
\toprule
\multicolumn{1}{c|}{\multirow{2}{*}{\textbf{Model}}} &
  \multicolumn{6}{c|}{\textbf{Annotated Explainability}} &
  \multicolumn{3}{c}{\textbf{Unannotated Explainability}} \\ \cmidrule{2-10} 
\multicolumn{1}{c|}{} &
  \textbf{BLEU} &
  \textbf{CIDEr} &
  \textbf{ROUGE-L} &
  \textbf{METEOR} &
  \textbf{SPICE} &
  \textbf{Avg.} &
  \textbf{Human-Eval} &
  \textbf{GPT4-Eval} &
  \textbf{Avg.} \\ \midrule
LLaVA-7B~\cite{liu2024visual} &
  0.183 &
  0.021 &
  0.139 &
  0.110 &
  0.085 &
  0.108 &
  2.368 &
  1.542 &
  1.955 \\
Llama3.2-Vision~\cite{dubey2024llama} &
  0.131 &
  0.009 &
  0.131 &
  0.081 &
  0.116 &
  0.093 &
  2.265 &
  1.667 &
  1.966 \\
Qwen2.5-VL-7B~\cite{wang2024qwen2} &
  0.140 &
  0.012 &
  0.143 &
  0.081 &
  0.150 &
  0.105 &
  2.034 &
  1.383 &
  1.709 \\
GPT-4o~\cite{achiam2023gpt} &
  0.123 &
  0.011 &
  0.082 &
  0.051 &
  0.072 &
  0.068 &
  2.559 &
  2.055 &
  2.307 \\ \midrule
Ours &
  \textbf{0.203} &
  \textbf{0.027} &
  \textbf{0.155} &
  \textbf{0.148} &
  \textbf{0.155} &
  \textbf{0.138} &
  \textbf{3.572} &
  \textbf{2.668} &
  \textbf{3.120} \\ \bottomrule
\end{tabular}
}
\vspace{-5mm}
\end{table*}

%% file: sec/5_conclusion.tex
\vspace{-1.5mm}
\section{Conclusion}

\vspace{-1.5mm}
In this paper, we propose $\mathcal{X}^2$-DFD, a \textbf{unified multimodal framework for \textit{explainable} and \textit{extendable} deepfake detection}.
For the first time, we systematically evaluate the intrinsic capabilities of the pre-trained MLLMs, revealing their varying effectiveness across different forgery-related features. 
Inspired by this, we implement a targeted fine-tuning strategy, which has largely improved the explainability of the MLLMs, specifically capitalizing on their strengths.
Furthermore, by integrating specific feature detectors (SFD), we design an adaptive fusion module to combine the complementary advantages of both MLLMs and conventional detectors for improved detection.
\vspace{-5pt}
\paragraph{Limitations and Future Work.} While our framework demonstrates strong performance in detecting identity-specific facial forgeries, it has certain limitations. \textbf{First}, exploration of entire face synthesis (EFS) forgeries remains relatively underdeveloped compared to face-swapping or attribute-editing techniques. Our work begins to address this gap, and future efforts should focus on deeper analysis and model improvements for EFS-generated content. \textbf{Second}, our current implementation focuses solely on static image detection. However, real-world applications increasingly involve multimodal forgeries across video and audio streams. Extending our method to handle videos and audio-visual deepfakes is a critical next step for building a comprehensive and practical detection system.
\vspace{-5pt}
\paragraph{Broader Impacts.} This research advances machine learning with a new framework to detect and explain Deepfake images, effectively identifying deepfakes and reducing misuse of generative models for significant societal benefit. However, it risks being used to improve deepfake realism. To counter this, following previous works~\cite{yan2024effort,yan2024df40} implement access controls. We will urge researchers to minimize harms while maximizing the positive impact of this work.
\vspace{-1.5mm}
\paragraph{Ethics \& Reproducibility statements.} 
All facial images used are from publicly available datasets with proper citations, ensuring no violation of personal privacy. And the human study has received the \textit{IRB approval}. 

\vspace{-3pt}


\vspace{-1.5mm}
\paragraph{Content Structure of the Appendix.}
Due to page constraints, we include additional analyses and experiments in the Appendix. Specifically, the Appendix contains the following sections: Overview of Appendix, Experiment Setting Details, Additional Experimental Results, Human Study and GPT4 Evaluation, Additional Analysis of Model Feature Assessment, Additional Analysis of Strong Feature Strengthening, Additional Analysis of Weak Feature Supplementing, Additional Analysis of Ablation Study, Sample Showing. 
\textbf{For further details, please refer to the Appendix.}